\pdfoutput=1

\documentclass[11pt]{article}

\usepackage{EACL2023}

\usepackage{graphicx}
\usepackage{amsmath}
\usepackage{mathtools}
\usepackage{amssymb}
\usepackage{times}
\usepackage{latexsym}
\usepackage{booktabs}
\usepackage{multirow}
\usepackage{adjustbox}
\usepackage{siunitx}
\usepackage{tabularx}

\usepackage{array}
\newcolumntype{P}[1]{>{\centering\arraybackslash}p{#1}}

\usepackage{algorithm}
\usepackage{algpseudocode}

\usepackage[T1]{fontenc}

\usepackage[utf8]{inputenc}

\usepackage{microtype}

\usepackage{array}
\newcolumntype{P}[1]{>{\centering\arraybackslash}p{#1}}
\newcolumntype{L}[1]{>{\raggedright\arraybackslash}p{#1}}

\newcolumntype{F}[1]{%
    >{\raggedright\arraybackslash\hspace{0pt}}p{#1}}%
\newcolumntype{T}[1]{%
    >{\centering\arraybackslash\hspace{0pt}}p{#1}}%

\title{MiniALBERT: Model Distillation via Parameter-Efficient Recursive Transformers}

\author{
  Mohammadmahdi Nouriborji$^{2\dagger}$, 
  Omid Rohanian$^{1,2\dagger}$,
  Samaneh Kouchaki$^{3}$,\\
  \textbf{David A. Clifton}$^{1,4}$\\
  $^1$Department of Engineering Science, University of Oxford, Oxford, UK \\
  $^2$NLPie Research, Oxford, UK \\
  $^3$Dept. Electrical and Electronic Engineering, University of Surrey, Guildford, UK \\
  $^4$Oxford-Suzhou Centre for Advanced Research, Suzhou, China \\
  \texttt{\{m.nouriborji,omid\}@nlpie.com}\\
  \texttt{samaneh.kouchaki@surrey.ac.uk}\\
  \texttt{\{omid.rohanian,david.clifton\}@eng.ox.ac.uk}\\
 }

\begin{document}
\maketitle
\def\thefootnote{$\dagger$}\footnotetext{The two authors contributed equally to this work.}
\def\thefootnote{\arabic{footnote}}

\begin{abstract}
Pre-trained Language Models (LMs) have become an integral part of Natural Language Processing (NLP) in recent years, due to their superior performance in downstream applications. In spite of this resounding success, the usability of LMs is constrained by computational and time complexity, along with their increasing size; an issue that has been referred to as `overparameterisation'. Different strategies have been proposed in the literature to alleviate these problems, with the aim to create effective compact models that nearly match the performance of their bloated counterparts with negligible performance losses. One of the most popular techniques in this area of research is model distillation. Another potent but underutilised technique is cross-layer parameter sharing. In this work, we combine these two strategies and present MiniALBERT, a technique for converting the knowledge of fully parameterised LMs (such as BERT) into a compact recursive student. In addition, we investigate the application of bottleneck adapters for layer-wise adaptation of our recursive student, and also explore the efficacy of adapter tuning for fine-tuning of compact models. We test our proposed models on a number of general and biomedical NLP tasks to demonstrate their viability and compare them with the state-of-the-art and other existing compact models. All the codes used in the experiments are available at \url{https://github.com/nlpie-research/MiniALBERT}.
Our pre-trained compact models can be accessed from \url{https://huggingface.co/nlpie}.

\end{abstract}

\section{Introduction}
\label{intro}

Following the introduction of BERT \citep{devlin-etal-2019-bert}, generic pre-trained  Language Models (LMs) have started to dominate the field of NLP. Virtually all state-of-the-art NLP models are built on top of some large pre-trained transformer as a backbone and are subsequently fine-tuned on their target dataset. While this pre-train and fine-tune approach has resulted in significant improvements across a wide range of NLP tasks, the widespread use of resource-exhaustive and overparameterised transformers has also raised concerns among researchers about their energy consumption, environmental impact, and ethical implications \citep{strubell-etal-2019-energy, bender2021dangers}.    

As a response to this, different approaches have appeared with the aim to make large LMs more efficient, accessible, and environmentally friendly. Model compression is a line of research that has recently received considerable attention. It involves encoding a larger and slower but more performant model into a smaller and faster one with the aim to retain much of the former's performance capability \citep{bucilua2006model}. Knowledge distillation \citep{hinton2015distilling}, quantisation \citep{shen2020q}, and pruning \citep{ganesh-etal-2021-compressing} are three examples of such methods. 

Adapter modules \citep{bapna-firat-2019-simple,he-etal-2021-effectiveness} are recently introduced as an effective mechanism to address the parameter inefficiency of large pre-trained models. In this approach, several `bottleneck adapters'\citep{houlsby2019parameter} are embedded inside different locations within the original network. During fine-tuning, the parameters of the original model are kept fixed, and for each new task only the adapters are fine-tuned. This only adds a small number of parameters to the overall architecture and allows for a much faster and more efficient fine-tuning on different downstream tasks. 

Another approach to improve efficiency of LM-based transformers is shared parameterisation, which was popularised by ALBERT \citep{lan2019albert}. While the original formulation of transformers \citep{vaswani2017attention} employs full parameterisation wherein each model parameter is independent of other modules and used only once in the forward pass, shared parameterisation allows different modules of the network to share parameters, resulting in a more efficient use of resources given the same parameterisation budget. However, a common downside of this approach is slower inference time and reduced performance. \citet{ge2022edgeformer} posits two different parameterisation methods as an attempt to address the compute and memory challenges of transformer models and explores layer-wise adaptation in an encoder-decoder architecture. These methods exploit cross-layer parameter sharing in a way that would allow for the model to be utilised on mobile devices with strict memory constraints while achieving state-of-the-art results on two seq2seq tasks for English.     

In this work, we exploit some of the above approaches to create a number of compact and efficient encoder-only models distilled from much larger language models. The contributions of this work are as follows:

\begin{itemize}
    \item To the best of our knowledge, we are the first to compress fully parameterised large language models using recursive transformers (i.e. ALBERT-like models that employ full parameter sharing). 
    
    \item We demonstrate the effectiveness of our pre-trained bottleneck adapters by merely fine-tuning them on downstream tasks while still achieving competitive results.
    
    \item We present several light-weight transformers with parameters ranging from $12$M for the smallest to $32$M for the largest. These models are shown to perform at the same level with their fully parameterised versions.
    
    \item Finally, we evaluate our models on a wide range of tasks and datasets on general and biomedial NLP datasets. 
\end{itemize}

\section{Background}
\label{background}

\begin{figure*}[ht!]
\centering
\includegraphics[scale=0.3]{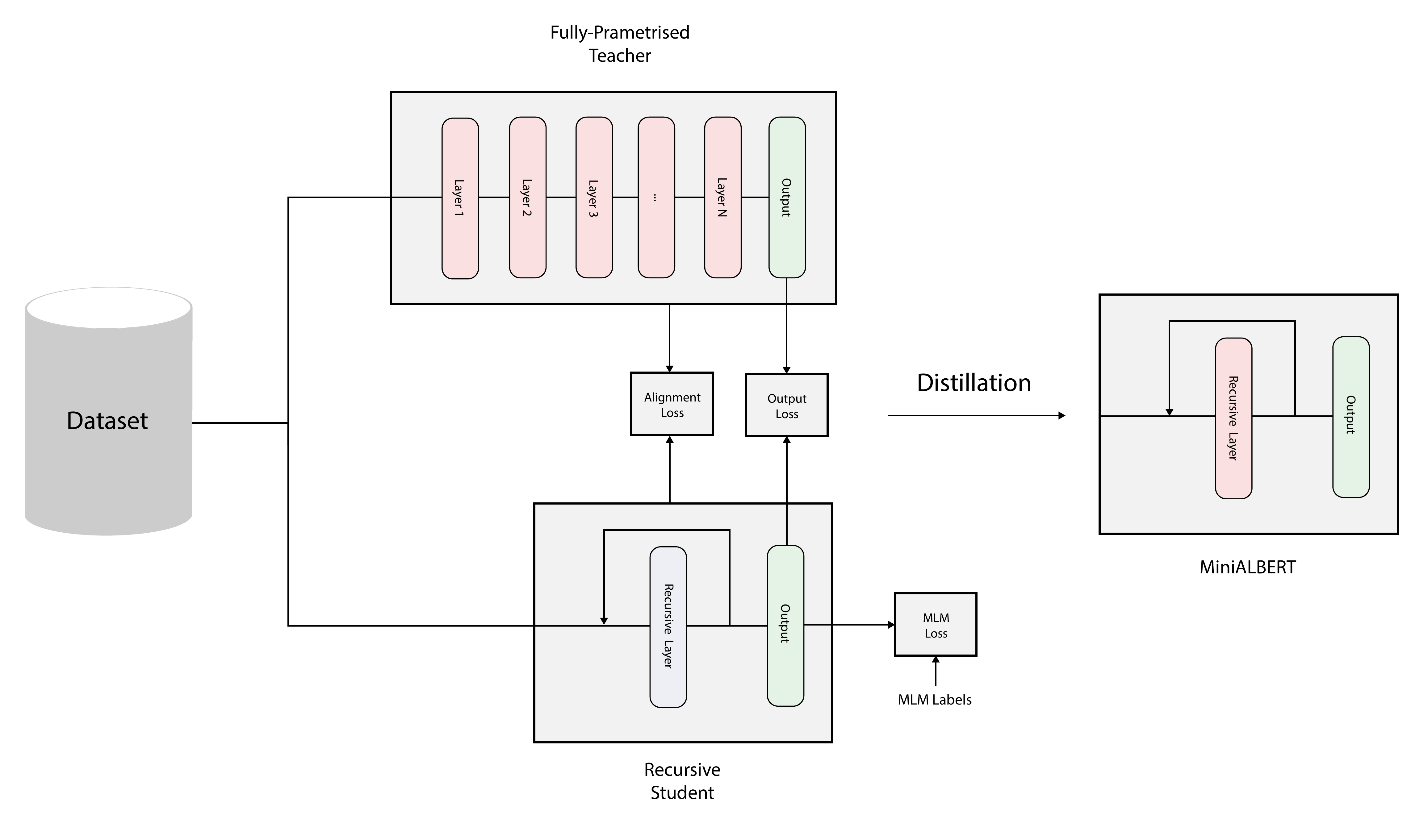}
\caption{The layer-to-layer distillation procedure proposed for distilling the knowledge of a fully-parameterised teacher into a compact recursive student. While the teacher has fully parameterised layers, the recursive student has only one layer and the output is fed back into the same layer repeatedly. Despite this compact structure, our proposed distillation procedure is designed to align the output of each iteration of the recursive student with a particular layer of the fully-parameterised  teacher, as if the student had fully-parameterised layers. Additional losses, namely, Output Loss, and MLM Loss, as shown above, are used for further knowledge distillation.}
\label{fig:distillation}
\end{figure*}

\subsection{LM-based Transformers and Computational Complexity}
\label{transformers}

Ever since the introduction of the transformer architecture \citep{vaswani2017attention}, large LM-based transformers such as BERT \citep{devlin-etal-2019-bert} have become increasingly more popular in NLP and lie at the heart of most state-of-the-art models. A transformer is primarily composed of a number of transformer blocks stacked on top of one another. BERT$_{Base}$ , for instance, consists of $12$ of these blocks. The most important component in a block is the multi-head self-attention module. To be useful for language tasks, transformers are pre-trained using a number of self-supervised auxiliary tasks \citep{xia-etal-2020-bert}; these usually include some variation of Language Modelling (LM) and an optional sentence-level prediction task. Examples of the former include Masked Language Modelling (MLM) and Casual Language Modelling (CLM). For the latter, BERT uses Next Sentence Prediction (NSP) and ALBERT \citep{lan2019albert} employs Sentence Order Prediction (SOP).  

The standard approach to utilise these pre-trained models is to fine-tune them on a target task. Given $N$ as the sequence length, the computational and time complexity of self-attention is $N^{2}$ \citep{keles2022computational}. In recent years, different approaches have appeared in the literature to address this bottleneck by modifying the self-attention operation in order to improve the general efficiency of transformers (with different performance trade-offs). \citet{tay2020efficient} surveys the most common approaches to develop what is referred to as `efficient transformers'.

The magnitude of the parameters of LM-based transformers is another significant issue that restricts their use. With new releases like GPT-3 and MT-NLG \citep{smith2022using} that feature hundreds of billions of parameters, these models have become increasingly overparameterised due to the large number of layers and embedding sizes \citep{rogers-etal-2020-primer}.

\subsection{Model Distillation}
\label{distillation}

The overparameterisation issue has motivated research into developing methods to compress larger models into smaller and faster versions that perform reasonably close to their larger counterparts. Knowledge distillation \citep{hinton2015distilling} is a prominent method that intended to distill a lightweight `student' model from a larger `teacher' network by using the outputs of the teacher netwrok as soft labels. Distillation can either be done task-specifically during fine-tuning, or task-agnostically by mimicking the MLM outputs or the intermediate representations of the teacher prior to the fine-tuning stage. The latter is more flexible and computationally less expensive \citep{wang2020minilm}. DistilBERT is a well-known example of a distilled model derived from BERT which is claimed to be $40\%$ smaller in terms of parameters and $60\%$ faster while retaining $97\%$ of BERT's performance on a range of language understanding tasks \citep{sanh2019distilbert}.

\subsection{Efficient Fine-tuning Approaches}
\label{efficient-tuning}

As discussed in Sec \ref{transformers}, LM-based transformers involve a large number of parameters and they are often fine-tuned on a target dataset. However, fine-tuning could become time-consuming as the size of the datasets grow. Different techniques exist in the literature to alleviate this bottleneck during fine-tuning. In this section we explore two of these techniques, namely, prompt tuning and bottleneck adapters. 

Prompt tuning \citep{lester-etal-2021-power} is a technique in which the weights of a language model are kept frozen during the fine-tuning stage and fine-tuning is reformulated as a cloze-style task. Similar to T5, prompt tuning regards all tasks as a variation of text generation and conditions the generation using `soft prompts'. A typical prompt consists of a text template with a masked token and a set of candidate label words to fill the mask. This turns the target task into another MLM objective in which the right candidates are chosen and soft prompts are learned. This method is especially useful for few-shot learning scenarios where there are not many target labels available for standard fine-tuning.  

Bottleneck Adapters (BAs) \citep{adapter,pfeiffer-etal-2021-adapterfusion,adapter_drop,adapter_hub} are another mechanism used during fine-tuning to enhance efficiency of training. Each BA block consists of a linear down-projection, non-linearity, and up-projection along with residual connections. Several of these adapters are placed after the feed-forward or attention modules in a transformer. Similar to prompts, only the BAs are trained during fine-tuning.         

\subsection{Parameter Sharing via Recursion}

 Weight sharing is a strategy intended to reduce the overall number of parameters in a model. \citet{lan2019albert} introduced cross-layer parameter sharing in a recurrent-like encoder-only architecture where instead of having several transformer blocks with different parameters, there is only one transformer block whose outputs are recursively fed into itself a number of times. This drastically reduces the number of unique parameters in the model.  

Edgeformer \citep{ge2022edgeformer} is a recent work towards development of parameter-efficient encoder-decoder models specialised for on-device seq2seq generation. EdgeFormer employs two novel approaches for cost-effective parameterisation which improve on standard cross-layer parameter sharing. In addition to efficient parameterisation, Edgeformer explores the use of layer-wise adaptation in encoder-decoder models.  To the best of our knowledge, however, the use of layer-wise adaptation in recursive encoder-only models (Sec \ref{sec:recursive-student}) is yet to be explored.

\section{Methods}
\label{methods}

In this work, we introduce a method to distil the knowledge of a fully parameterised transformer into a lightweight efficient recursive transformer via layer-to-layer distillation. We also experiment with layer-wise adaption of the recursive transformer via bottleneck adapters and factorise the embedding layer for extra parameter saving. In this section, we explain each component of our compact models in detail.  

\subsection{The Fully Parameterised Transformer}

For each layer $i$, let the multi-head attention and feed-forward blocks be $f_{att}^{i}(x)$ and $f_{mlp}^{i}(x)$, respectively. The output of each layer is computed as follows: 

\begin{equation} \label{eq:1}
    O_i = f_{mlp}^{i}(f_{att}^{i}(O_{i-1}))
\end{equation}
where $O_{i}$ is the output of the $i^{th}$ layer of the transformer.

\subsection{The Recursive Student}
\label{sec:recursive-student}

The output of the recursive student is computed as follows:
\begin{equation} \label{eq:2}
    O_i = f_{mlp}(f_{att}(O_{i-1}))
\end{equation}
where $O_{i}$ is the output of the $i^{th}$ iteration of the recursive transformer. Note that unlike Equation \ref{eq:1}, the multi-head attention and feed-forward blocks in this case are layer-agnostic, i.e. the same layers are used recursively.

\begin{figure}[ht!]
\centering
\includegraphics[scale=0.4]{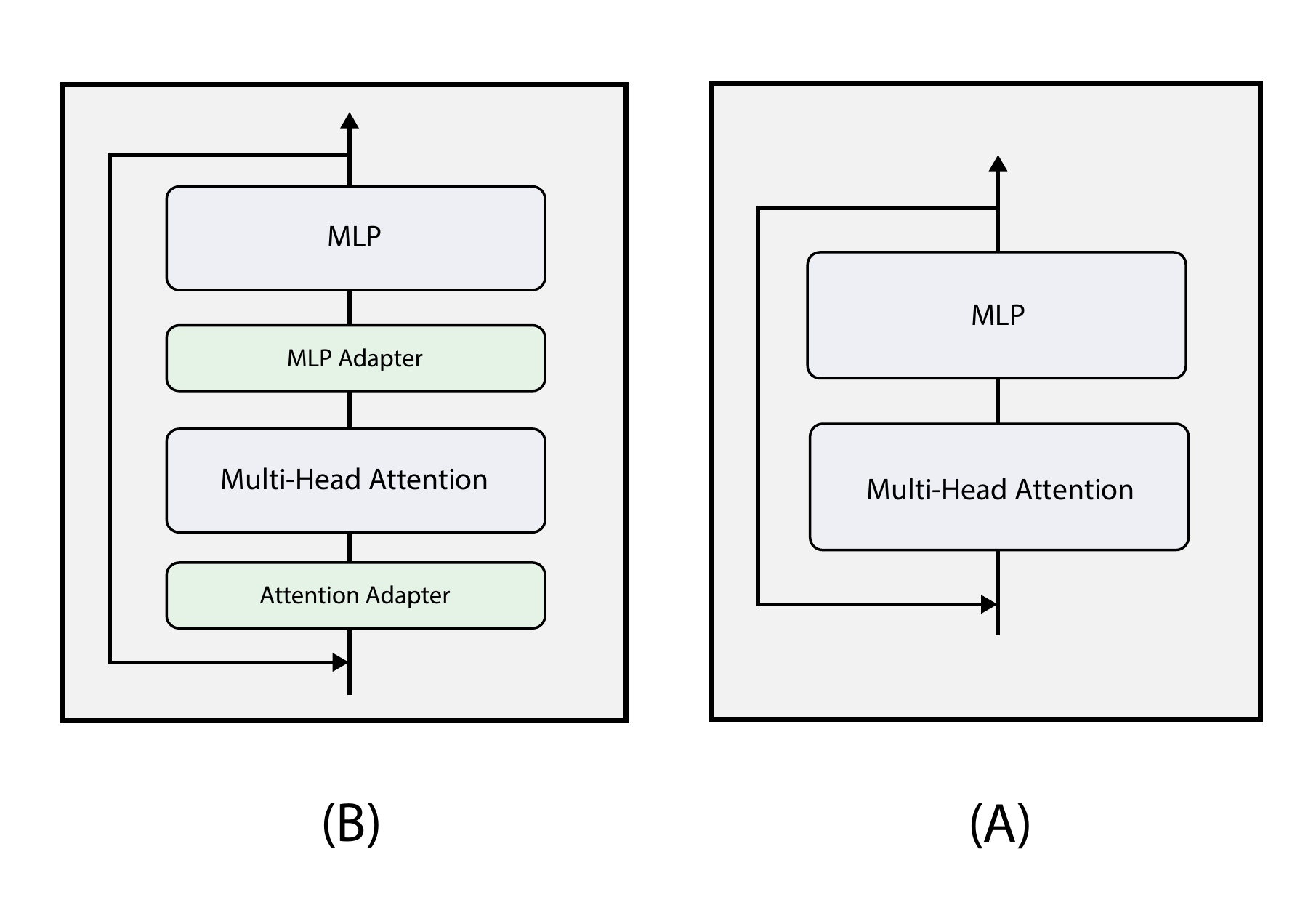}
\caption{The two recursive students proposed in this work. (A) is a simple recursive student that employs cross-layer parameter sharing, which means that the multi-head attention and MLP blocks are shared across all layers. (B) is a recursive student with cross-layer parameter sharing which additionally uses layer-wise adaption via bottleneck adapters.}
\label{fig:architecture}
\end{figure}

\subsection{Recursive Student with Layer-wise Adaption}
This formulation is identical to the recursive student defined in Section \ref{sec:recursive-student}, except that, in addition to the shared parameters, here we employ a small number of trainable parameters. This will allow the model to capture distinct features in each iteration, similar to how transformer layers behave in a fully parameterised model. To this end, we use ‌‌‌‌Bottleneck Adapters (BAs) which are small bottleneck blocks followed by residual connections, as defined in the following equation:
\begin{equation}
\phi(X) = W_{up}\ \ \sigma(W_{down}\ \ X) + X
\end{equation}
where $\phi(.)$ represents the BA and $\sigma(.)$ is a non-linearity function such as \textit{RELU} or \textit{GELU}. The recursive student with layer-wise adaption can be formulated as follows:
\begin{equation}
    O_{i} = f_{mlp}(\phi_{mlp}^{i}(f_{att}(\phi_{att}^{i}(O_{i-1}))))
\end{equation}
where $\phi_{att}^{i}(.)$ and $\phi_{mlp}^{i}(.)$ are the BAs for the multi-head attention and feed-forward blocks of the $i^{th}$ iteration of the recursive transformer.

\subsection{Embedding Factorisation}
Following the work of ALBERT \citep{lan2019albert}, instead of a full-rank embedding matrix, we use a low-rank matrix with size $|V| \times r$ where $V$ is the vocabulary  and $r$ is the rank of the embedding matrix. We additionally use a projection weight with the size of $r \times d$ where $d$ is the hidden dimension of the transformer. This is set to $768$ in our experiments. Factorisation can be mathematically expressed as

\begin{equation} \label{eq_factorization}
E = E_{low}\  W_{e}
\end{equation}
where $E_{low} \in \mathbb{R}^{|V| \times r}$, $W_{e} \in \mathbb{R}^{r \times d}$ and $E \in \mathbb{R}^{|V| \times d}$. In our experiments, we employ factorisation with ranks $312$ and $128$, and explore initialising the $312$ versions with pre-trained embeddings obtained from the TinyBERT \citep{jiao2019tinybert} for the general models and TinyBioBERT \citep{rohanian2022effectiveness} for the biomedical models.

In our experiments, we employed factorisation with ranks of $312$ and $128$ and initialised the 312 versions with pre-trained embeddings from TinyBERT \citep{jiao2019tinybert} for general models and TinyBioBERT \citep{rohanian2022effectiveness} for biomedical models.


\subsection{Distillation Procedure}
We use three different loss terms, namely, MLM, alignment, and output. The MLM loss, which is the original loss used in Masked Language Modelling, is defined as follows:
\begin{equation}
    L_{MLM}(X, Y) = \sum_{n=1}^{N}CE(f_{s}(X)^{n}, Y^{n})
\end{equation}
where $X$ denotes the the model's input, $N$ the number of input tokens, $CE$ the cross-entropy loss function, {$Y^{n}$} the one-hot encoded label for the $n^{th}$ token\footnote{ $Y^{n}$ is a zero vector if the $n^{th}$ token is not masked.}, and $f_{s}(X)^{n}$ the output distribution of the student for the $n^{th}$ token. $Y^{n}$ and $f_{s}(X)^{n}$ are both $|V|$-dimensional.

Because the number of layers in the student network is less than that of the teacher, alignment is typically achieved by comparing the student's layers to a subset of the teacher's. However, in our case, the student is recursive and lacks multiple layers. Therefore, in each iteration, we map the output of the student's layer to a specific layer of the teacher, i.e. the first iteration is considered the first layer, the second iteration is the second layer, and so on.  The alignment loss is a combination of two terms, namely, attention and hidden losses. Attention loss is used to align the student's attention maps with those of the teacher and is defined as follows: 
\begin{equation} 
    L_{att}(\hat{A}, A) = \frac{1}{HN}\sum^H_{h=1} \sum^{N}_{n=1} D_{KL} (\hat{A}^{h}_{n} \hspace{2.5pt} || \hspace{2.5pt} A^{h}_{n}) 
\end{equation}
$\hat{A}$ and $A$ are the inputs to the loss function, and they correspond to the attention maps of a certain layer of the student and its associated layer of the teacher, respectively.
The cosine-based hidden loss is used to align the hidden states of the student and teacher, and is defined as:
\begin{equation} 
    L_{hidden}(\hat{H}, H) = \frac{1}{N}\sum^N_{n=1} 1 - \psi(\hat{H}^{n}, H^{n})
\end{equation}
where $\hat{H}$ and $H$ are the hidden states of a particular layer in the student and teacher networks. The $\psi$ function denotes cosine similarity\footnote{$\psi(\Vec{u}, \Vec{v}) = \frac{\Vec{u}.\Vec{v}}{||\Vec{u}||_{2}||\Vec{v}||_{2}}$}. The alignment loss is defined as:
\begin{align}
    \label{eq:align}
    L_{align}(\hat{A},A,\hat{H},H) = & \hspace{15pt} L_{att}(\hat{A}, A) \\
                                     & + L_{hidden}(\hat{H}, H) \nonumber  
\end{align}

The output loss is based on KL divergence and is inteneded to align the output distribution of the student with the teacher on the MLM objective. This loss term is defined as below:  
\begin{equation}
L_{out}(X) = \sum_{n=1}^{N}W_{n}D_{KL}(f_{s}(X)^n \hspace{2.5pt} || \hspace{2.5pt} f_{t}(X)^n)    
\end{equation}
where $f_{s}(X)^n$ and $f_{t}(X)^n$ are the output distributions of the student and teacher for the $n^{th}$ token, respectively. $W_{n}$ is $1$ if the $n^{th}$ token is masked and $0$ otherwise. This ensures that only the masked tokens will contribute to the loss. 


The complete layer-to-layer distillation loss used in this study is expressed by the following equation: 

\begin{align}
\label{eq:full-los}
    L(X, Y, &A_{s}, A_{t}, H_{s}, H_{t}) = \\ \nonumber
              & \hspace{14pt}\lambda_{1}\ L_{MLM}(X, Y)\\ 
              & + \lambda_{2}\sum_{l=1}^{L}L_{align}(A_{s}^{l}, A_{t}^{g(l)}, H_{s}^{l}, H_{t}^{g(l)}) \nonumber \\
              & + \lambda_{3}\ L_{out}(X) \nonumber.
\end{align}

$A_{s}$ and $A_{t}$ in Equation \ref{eq:full-los} are collections of attention maps for the student and teacher, respectively. $H_{s}$ and $H_{t}$ are sets of hidden states for the student and teacher. $L$ is the number of iterations (layers) of the recursive student. $g(.)$ is a mapping function that connects each iteration of the student to a specific layer of the teacher. Finally, $\lambda_{1}$ to $\lambda_{3}$ are hyperparameters used for controlling the importance of each component of the loss function (we use $\lambda_{1}=1.0$, $\lambda_{2}=3.0$, $\lambda_{3}=5.0$ in our experiments).

\begin{table*}[ht!]
    \centering
    \caption{\label{t:glue} The DEV set results on the GLUE benchmark. ALBERT$_{6}$ and ALBERT$_{12}$ denote ALBERT models with $6$ and $12$ layers respectively, and an embedding size of $128$, which are trained on the same data for the same number of training steps as MiniALBERT. DistilBERT$_{base}$ is a DistilBERT model trained with the same distillation setting as \citet{sanh2019distilbert} and for the same number of training steps as MiniALBERT. Adapter denotes layer-wise adaption and EF denotes Embedding Factorisation. The metrics used for reporting the results on each dataset is the same as the official GLUE benchmark. $\dagger$ denotes that the models were trained using adapter tuning in which all weights of the model except bottleneck adapters are kept frozen during fine-tuning. $*$ shows that the bottleneck adapters were initialised randomly since the model has not used bottleneck adapters at the pre-training stage. N/A means that the model did not learn anything from the target dataset. \#Params denotes the number of tunable parameters which are used during fine-tuning. Note that for fine-tuning TinyBERT on the downstream tasks, unlike the original paper \citep{jiao2019tinybert}, we do not employ task-specific distillation; this is to ensure the comparison with other models is fair.}
    \scalebox{0.7}{
    \begin{tabular}{L{3cm}P{1cm}P{0.5cm}P{1.5cm}P{2.5cm}P{1cm}P{1cm}P{1cm}P{1cm}P{2cm}P{1cm}P{1cm}P{1cm}}
        \toprule[1pt]
        Model & Adapter & EF & \#Params & MNLI-(m/mm) & QQP & QNLI & SST-2 & CoLA & STS-B & MRPC & RTE & Avg  \\\midrule[0.5pt]
        BERT-base & -  & - & 110M & \textbf{84.78}/\textbf{84.78} & \textbf{87.91} & \textbf{91.56} & \textbf{92.77} & \textbf{57.19} & \textbf{89.20} & \textbf{91.46} & \textbf{74.72} & \textbf{83.81}\\
        DistilBERT & - & - & 65M & 82.17/82.33 & 87.08 & 89.47 & 90.25 & \underline{53.61} & 87.20 & 88.66 & 66.42 & 80.79\\
        MobileBERT & - & - & 25M & \underline{84.03}/\underline{83.84} & \underline{87.41} & \underline{91.17} & \underline{91.16} & 51.55 & \underline{88.12} & \underline{90.59} & \underline{66.78} & \underline{81.62}\\
        TinyBERT & - & - & 15M & 80.22/80.65 & 84.44 & 85.31 & 88.87 & 17.53 & 87.77 & 89.41 & 66.06 & 75.58\\
        \cmidrule[0.5pt]{1-13}

        DistilBERT$_{base}$ & - & - & 65M & \textbf{82.08}/\textbf{82.27} & \textbf{86.01} & \underline{86.91} & \textbf{91.28} & \textbf{50.33} & \underline{83.92} & \underline{88.06} & \underline{59.56} & \textbf{78.93}\\
        ALBERT$_{6}$ & - & $\checkmark$ & 11M & 76.35/77.01 & 84.73 & 84.51 & 86.12 & 29.54 & 82.76 & 86.39 & \underline{59.56} & 74.10\\
        ALBERT$_{12}$ & - & $\checkmark$ & 11M & \underline{78.93}/\underline{80.03} & \underline{85.45} & \textbf{88.12} & \underline{86.92} & \underline{41.73} & \textbf{86.21} & \textbf{90.65} & \textbf{63.17} & \underline{77.91}\\
        
        \cmidrule[0.5pt]{1-13}
        
        MiniALBERT$_{768}$ & $\times$ & $\times$ & 31M & 80.71/\underline{81.80} & 86.25 & 88.13 & \underline{89.79} & 43.71 & \textbf{86.93} & 89.00 & 61.37 & 78.63\\
        MiniALBERT$_{312}$ & $\times$ & $\checkmark$ & 17M & 80.55/81.29 & \underline{86.46} & 87.46 & 89.56 & 45.82 & 85.51 & \underline{89.22} & 62.09 & 78.66\\
        MiniALBERT$_{768}$ & $\checkmark$ & $\times$ & 32M & \textbf{81.04}/\textbf{82.05} & \textbf{86.48} & \textbf{88.79} & \underline{89.79} & \underline{46.33} & \underline{86.92} & 88.46 & 67.50 & 79.70\\
        MiniALBERT$_{312}$ & $\checkmark$ & $\checkmark$ & 18M & \underline{80.78}/81.67 & 86.35 & \underline{88.57} & \textbf{90.36} & 45.50 & 86.60 & 88.96 & \textbf{69.31} & \underline{79.78}\\
        MiniALBERT$_{128}$ & $\checkmark$ & $\checkmark$ & 12M & 80.64/81.33 & 86.29 & 88.08 & 89.67 & \textbf{47.93} & 86.62 & \textbf{89.77} & \underline{68.95} & \textbf{79.92}\\
        
        \cmidrule[0.5pt]{1-13}
        
        MiniALBERT$_{768}$$^{\dagger,*}$ & $\times$ & $\times$ & 0.9M & 74.61/75.72 & 79.72 & 80.70 & 85.20 & N/A & 83.90 & 81.72 & 52.70 & 68.25\\
        MiniALBERT$_{312}$$^{\dagger,*}$ & $\times$ & $\checkmark$ & 0.9M & 74.48/75.70 & 79.76 & 82.39 & 83.48 & N/A & 80.94 & 81.22 & 54.15 & 68.01\\
        
        MiniALBERT$_{768}$$^{\dagger}$ & $\checkmark$ & $\times$ & 0.9M & \textbf{79.48}/\underline{80.06} & \textbf{85.29} & \textbf{87.84} & \underline{90.02} & 42.26 & \textbf{86.33} & 87.74 & \textbf{67.14} & \textbf{78.46}\\
        MiniALBERT$_{312}$$^{\dagger}$ & $\checkmark$ & $\checkmark$ & 0.9M & \underline{79.13}/\textbf{80.16} & \underline{85.27} & 86.10 & 89.90 & \textbf{45.25} & \underline{85.34} & \underline{87.91} & 65.34 & 78.26\\
        MiniALBERT$_{128}$$^{\dagger}$ & $\checkmark$ & $\checkmark$ & 0.9M & 78.05/79.66 & 84.94 & \underline{87.40} & \textbf{90.36} & \underline{44.72} & 84.62 & \textbf{89.08} & \underline{66.78} & \underline{78.40}\\
        
        \bottomrule
    \end{tabular}}
    \vspace{10pt}
\end{table*}

\section{Experiments and Results}
\label{res}
We evaluate our general models on the widely used GLUE benchmark \citep{wang2018glue}. All the models were pre-trained on four Tesla V$100$ $32$GB GPUs with a total batch size of $192$ ($48$ each) and fine-tuning was done using only one Tesla V100. A random seed of $42$ was used consistently throughout training for fair comparison. For all of the datasets, in order to do full fine-tuning, we use a learning rate from the set \{$5$e-$5$, $3$e-$5$, $1$e-$5$\}. For large datasets (MNLI, QQP, QNLI, and SST-2), we train models for a maximum of $5$ epochs, and up to ten epochs on other datasets. The hyperparameters used for full-finetuning are listed in Table \ref{t:glue-ft}. 

For adapter-tuning, the learning rate was selected from the set \{$5$e-$5$, $5$e-$4$, $1$e-$3$\}. Models were trained for a maximum of $10$ epochs for large datasets and up to $15$ epochs for other datasets. Table \ref{t:glue-at} details the hyperparameters used during adapter tuning on the GLUE benchmark. The results of the baselines and our general models are available in Table \ref{t:glue}.

The biomedical models are evaluated on the task of Named Entity Recognition (NER), which is one of the most prominent tasks in biomedical NLP. We use four well-known datasets, namely, NCBI-disease \citep{dougan2014ncbi}, BC5CDR-disease \citep{li2016biocreative}, BC5CDR-chem \citep{li2016biocreative}, and BC2GM \citep{smith2008overview}.

We generally follow the same pre-processing pipeline as \citet{rohanian2022effectiveness}. For biomedical NER, we use pre-processed datasets from \citet{lee2020biobert}. We fine-tune the models using a learning rate from the set \{$5$e-$5$, $3$e-$5$, $1$e-$5$\} and perform adapter-tuning with a learning rate from \{$5$e-$5$, $5$e-$4$, $1$e-$3$\}. The hyperparameters for the biomedical datasets are presented in Tables \ref{t:bio-ft} and \ref{t:bio-at}. The results of the baselines and our biomedical models are available in Table \ref{t:biomedNER}.

\begin{table*}[ht!]
    \centering
    \caption{\label{t:biomedNER} The results of biomedical NER for the biomedical baselines and our models distilled from BioBERT-v1.1 on the PubMed dataset. BioALBERT$_{6}$ and BioALBERT$_{12}$ represent ALBERT models with $6$ and $12$ layers, respectively, and an embedding size of $128$. These models were trained for the same number of steps as BioMiniALBERT using the same data. ``Adapter'' refers to layer-wise adaptation and ``EF'' stands for Embedding Factorization. The results are reported using the F1-score as the evaluation metric. The notations here are consistent with Table \ref{t:glue}.}
    \scalebox{0.7}{
    \begin{tabular}{L{3cm}P{1cm}P{0.5cm}P{1cm}P{3cm}P{3cm}P{2.5cm}P{2.5cm}P{2.5cm}}
        \toprule[1pt]
        Model & Adapter & EF & \#Params & NCBI-disease & BC5CDR-disease & BC5CDR-chem & BC2GM & Avg  \\\midrule[0.5pt]
        DistilBioBERT & - & - & 65M & 87.93 & \underline{85.42} & \underline{94.53} & 86.60 & 88.62\\
        CompactBioBERT & - & - & 65M & \textbf{88.67} & 85.38 & 94.31 & \underline{86.71} & \underline{88.76}\\
        TinyBioBERT & - & - & 15M & 85.22 & 81.28 & 92.20 & 82.52 & 85.30\\
        BioMobileBERT & - & - & 25M & 87.21 & 84.62 & 94.23 & 85.26 & 87.83\\
        BioBERT & - & - & 110M & \underline{88.62} & \textbf{86.67} & \textbf{94.73} & \textbf{87.62} & \textbf{89.41}\\
        
        \cmidrule[0.5pt]{1-9}
        BioALBERT$_{6}$ & - & $\checkmark$ & 11M & \textbf{86.07} & \textbf{82.00} & \textbf{93.19} & \textbf{84.51} & \textbf{86.44}\\
        BioALBERT$_{12}$ & - & $\checkmark$ & 11M & \textbf{86.07} & \underline{81.94} & \underline{93.11} & \underline{84.33} & \underline{86.36}\\
        
        \cmidrule[0.5pt]{1-9}
        BioMiniALBERT$_{768}$ & $\times$ & $\times$ & 31M & 87.44 & 84.40 & 94.18 & 86.06 & 88.02\\
        BioMiniALBERT$_{312}$ & $\times$ & $\checkmark$ & 17M & 87.94 & 84.45 & 94.03 & 86.03 & 88.11\\
        BioMiniALBERT$_{768}$ & $\checkmark$ & $\times$ & 32M & \underline{88.02} & \textbf{84.98} & \textbf{94.49} & \underline{86.10} & \textbf{88.39}\\
        BioMiniALBERT$_{312}$ & $\checkmark$ & $\checkmark$ & 18M & \textbf{88.03} & \underline{84.75} & \underline{94.23} & \textbf{86.14} & \underline{88.28}\\
        BioMiniALBERT$_{128}$ & $\checkmark$ & $\checkmark$ & 12M & 87.16 & 84.58 & 94.20 & 86.00 & 87.98\\
        
        \cmidrule[0.5pt]{1-9}
        BioMiniALBERT$_{768}$$^{\dagger,*}$ & $\times$ & $\times$ & 0.9M & 85.61 & 82.31 & 92.80 & 84.89 & 86.40\\
        BioMiniALBERT$_{312}$$^{\dagger,*}$ & $\times$ & $\checkmark$ & 0.9M & 85.98 &  81.72 & 91.99 & 84.65 & 86.08\\
        BioMiniALBERT$_{768}$$^{\dagger}$ & $\checkmark$ & $\times$ & 0.9M & \textbf{87.80} & \textbf{84.64} & \underline{94.20} & \textbf{86.02} & \textbf{88.16}\\
        BioMiniALBERT$_{312}$$^{\dagger}$ & $\checkmark$ & $\checkmark$ & 0.9M & 87.61 & \underline{84.55} & 94.12 & 85.60 & 87.97\\
        BioMiniALBERT$_{128}$$^{\dagger}$ & $\checkmark$ & $\checkmark$ & 0.9M & \underline{87.71} & 84.48 & \textbf{94.22} & \underline{85.87} & \underline{88.07}\\
        \bottomrule
    \end{tabular}}
    \vspace{10pt}
\end{table*}

\section{Discussion}
\label{disc}

We trained our recursive students both with and without adapters and found that generally having adapters would increase the learning capacity of the model since each iteration provides an extra degree of freedom to the model in order to capture a specific type of feature. As shown in Tables \ref{t:glue} and \ref{t:biomedNER}, in virtually all of the studies, models with adapters outperformed models of comparable size without adapters. In general, both models with and without adapter have outperformed their baseline by a significant margin (as shown in Tables \ref{t:biomedNER} and \ref{t:glue}),  demonstrating the effectiveness of the proposed layer-to-layer distillation loss for constructing powerful compact recursive models. Comparison between the attention maps of our trained student and teacher models (Figure \ref{fig:attention} in appendix) suggests the specific recursive architecture we have introduced in this work is indeed capable of mimicking the components of a larger model. 

In addition, our experiments revealed that utilising adapters in the pre-training stage enables the model to effectively use adapter-tuning with only minor performance drops. However, adapter-tuning in models that have not used adapters in the pre-training stage causes major performance drops. We also found that adapter tuning is nearly $30\%$ faster than full fine-tuning. Therefore, with adapter-tuning, the models can be trained for a larger number of epochs given the same training time, which can potentially increase the performance of the model. In general, a higher learning rate was required for adapter-tuning than for full fine-tuning.  For both general and biomedical tasks we found that a learning rate of $5e-4$ results in the best performance, however, in some cases, a higher or lower learning rate such as $5e-5$ or $1e-3$ was deemed better. 

Another method explored in this work for parameter saving is Embedding Factorisation. In our experiments, we observed that regardless of drastic parameter reduction, models using this approach are still able to perform on par or even in some cases better than models with full-rank embedding. 


\section{Ablation Studies}
\subsection{The Effect of The Alignment Loss}
One of the main losses explored in this work for knowledge distillation is the alignment loss as explained in Equation \ref{eq:align}. Alignment loss is used for mapping each iteration of the recursive student to a specific layer of the teacher, so the knowledge of each fully-parameterised layer is explicitly encoded into a specific iteration of the recursive student. The alignment loss consists of two losses, one for aligning the attention maps and one for aligning the hidden states. For ablation studies, we trained our best model for $20$k steps on the Wikipedia dataset, with different alignment losses, and evaluated the resulting models on the GLUE dataset. The results are shown in Table \ref{t:alignment}.

\begin{table}[h!]
    \centering
    \caption{\label{t:alignment} Ablation study on the alignment loss. The performance drop is computed based on the difference with the average score of the model with full alignment (i.e. when all the alignment losses are used).}
    \begin{tabularx}{0.45\textwidth}{
    >{\raggedright\arraybackslash}X 
    >{\centering\arraybackslash}X 
    }
        \toprule[1pt]
        Alignment Type & Performance Drop\\\midrule[0.5pt]
        Hidden-Only & -0.5\\
        Attention-Only & -3.27\\
        No-Alignment & -5.27\\
        \bottomrule
    \end{tabularx}
\end{table}

As Shown in Table \ref{t:alignment}, without alignment, the performance of the recursive student drops significantly which shows the importance of using a layer-to-layer distillation technique. Furthermore, we discovered that aligning hidden states is far more important than aligning attention maps. We even noticed occasional improvements in some tasks by solely aligning hidden states, but the average performance of full alignment remains higher than hidden states alignment. 


\subsection{The Effect of an Extra Embedding Loss}
Following the work of \citet{jiao2019tinybert} and as part of our ablation tests, we investigated employing an extra loss for aligning the embeddings of the recursive student and fully-parameterised teacher.
This embedding loss is defined as follows: 
\begin{equation} \label{eq:embed_loss}
    L_{embed}(\hat{E}, E) = \frac{1}{N}\sum^N_{n=1} 1 - \psi(\hat{E}^n, E^n)
\end{equation}

$\hat{E}$ and $E$ are the inputs to the loss function and represent the embeddings of the student and teacher. The embedding weights are not aligned globally in this formulation; instead, the local embeddings created for each training sample are compared before entering the transformer encoder. For our ablation studies, we trained models on the Wikipedia dataset for $20$k steps, with and without this extra alignment loss, and evaluated them on the GLUE benchmark.The average performance of the models are reported in Table \ref{t:embedding}.

\begin{table}[h!]
    \centering
    \caption{\label{t:embedding} Ablation study on the embedding loss. `Full-rank' denotes models without embedding factorisation, and `Loss' denotes extra embedding loss during distillation.}
    \begin{tabularx}{0.45\textwidth}{
    >{\raggedright\arraybackslash}X 
    >{\centering\arraybackslash}X 
    }
        \toprule[1pt]
        Model & Avg Performance\\\midrule[0.5pt]
        Full-rank & 76.01\\
        Full-rank + Loss & 75.24\\
        Factorised & 76.31\\
        Factorised + Loss & 76.14\\
        \bottomrule
    \end{tabularx}
\end{table}

We discovered that, unlike the trend seen in fully-parameterised models, embedding loss reduces the performance of the student model (Table \ref{t:embedding}).
This difference between the recursive student and the fully parameterised teacher implies that forcing the recursive student's embedding to match that of the fully parameterised teacher  may not be beneficial, and that allowing the student to learn embeddings independently works better.



\section{Conclusion and Future Works}
\label{concl}

In this work, we explored distilling the knowledge of fully-parameterised language models into recursive students with cross-layer parameter sharing. We used a layer-to-layer distillation technique to observe the learning capacity of the recursive students compared to their fully-parameterised teachers. We used bottleneck adapters for improving the performance of our recursive students and also assessed the benefits of adapter-tuning at the fine-tuning stage. Furthermore, an embedding factorisation technique was used for additional parameter reduction, which was evaluated with and without an extra distillation loss to match the student's embeddings with the teacher. Finally, by integrating all of the strategies outlined above, we were able to train compact recursive students with no more than $12$M parameters, yielding competitive performance on both general and biomedical NLP. In the future, we hope to investigate various parameter-sharing and embedding factorisation strategies, as well as other layer-wise adaption techniques such as prompt-tuning. We would also like to train recursive students with larger hidden sizes and more training iterations to compress massive fully-parameterised models with minor performance drops. 

\section*{Funding}
This work was supported in part by the National Institute for Health Research (NIHR) Oxford Biomedical Research Centre (BRC), and in part by an InnoHK Project at the Hong Kong Centre for Cerebro-cardiovascular Health Engineering (COCHE). OR acknowledges the support of the Medical Research Council (grant number MR/W01761X/). DAC was supported by an NIHR Research Professorship, an RAEng Research Chair, COCHE, and the Pandemic Sciences Institute at the University of Oxford. The views expressed are those of the authors and not necessarily those of the NHS, NIHR, MRC, COCHE, or the University of Oxford.

\section*{Limitations}
Regardless of the parameter reduction induced by our proposed recursive architecture, the resultant models have the same inference latency and memory complexity as fully-parameterised models of comparable size, which for our models is DistilBERT. 

In general, our models' capacity for learning semantic and grammatical knowledge is limited, and they may perform poorly on tasks that necessitate a significant amount of reasoning and understanding, such as Question Answering or Semantic Acceptability. More analysis is required to determine the source of this limitation, i.e. whether it is a result of the architecture used or a consequence of the particular cross-layer sharing method etc.

\section*{Ethics Statement}
In this study, we aimed to create efficient lightweight versions of large and less accessible NLP models.
This area of research aims to make AI/NLP models more readily available, with fewer computational resources required to run them and potentially less negative environmental impact.

This work does not use any private or sensitive data and instead relies on widely used publicly available datasets that have been utilised by other researchers in the field with references provided in the paper for more information. All the codes and models are going to be made available for reproducibility purposes.


\bibliography{anthology,custom}
\bibliographystyle{acl_natbib}

\appendix
\section{Details of the Pre-Training}
\subsection{General Models}
For pre-training our general models we use the English subset of the Wikipedia dataset which is available on the Huggingface platform. For pre-processing the Wikipedia dataset, we use the `bert-base-uncased' tokeniser and apply a sliding window with a size of $256$ tokens and a stride size of $128$. Due to computational restrictions, we limit the maximum number of tokenised samples per article to $10$, resulting in a total of $21$ million training samples of $256$ tokens each. We then follow BERT's masking approach, with a masking probability of $15\%$ for each token. 

\begin{table}[h!]
    \centering
    \caption{\label{t:general-pretraining} Hyperparameters used for pre-training models on Wikipedia dataset}
    \begin{tabularx}{0.45\textwidth}{
    >{\raggedright\arraybackslash}X 
    >{\centering\arraybackslash}X 
    }
        \toprule[1pt]
        Param & Value\\\midrule[0.5pt]
        learning rate & $5e-4$\\
        scheduler & Linear\\
        optimiser & AdamW\\
        weight decay & $1e-4$\\
        total batch size & $192$\\
        warmup steps & $5000$\\
        epochs & $1$\\
        \bottomrule\\
    \end{tabularx}
\end{table}

\subsection{Biomedical Models}
We pre-trained the biomedical models using the PubMed Abstracts dataset which consists of $31$ million abstracts from PubMed articles. In the pre-processing stage, we employed the `bert-base-cased' tokeniser with a maximum length of $256$ and adhered to the same masking strategy as used in the training of the general models.

\begin{table}[h!]
    \centering
    \caption{\label{t:bio-pretraining} Hyperparameters used for pre-training models on Wikipedia dataset}
    \begin{tabularx}{0.45\textwidth}{
    >{\raggedright\arraybackslash}X 
    >{\centering\arraybackslash}X 
    }
        \toprule[1pt]
        Param & Value\\\midrule[0.5pt]
        learning rate & $5e-4$\\
        scheduler & Linear\\
        optimiser & AdamW\\
        weight decay & $1e-4$\\
        total batch size & $192$\\
        warmup steps & $5000$\\
        training steps & $100$K\\
        \bottomrule\\
    \end{tabularx}
\end{table}

\section{Finetuning Details}

\begin{table}[h!]
    \centering
    \caption{\label{t:glue-ft} Hyperparameters used for full fine-tuning of the models on the GLUE benchmark}
    \begin{tabularx}{0.45\textwidth}{
    >{\raggedright\arraybackslash}X 
    >{\centering\arraybackslash}X 
    }
        \toprule[1pt]
        Param & Value\\\midrule[0.5pt]
        learning rate & \{$5$e-$5$, $3$e-$5$, $1$e-$5$\}\\
        scheduler & Linear\\
        optimiser & AdamW\\
        weight decay & $0.01$\\
        batch size & \{$8$, $16$, $32$\}\\
        epochs & \{$3$, $5$, $10$\}\\
        \bottomrule\\
    \end{tabularx}
\end{table}

\begin{table}[h!]
    \centering
    \caption{\label{t:glue-at} Hyperparameters used for adapter-tuning of the models on the GLUE benchmark}
    \begin{tabularx}{0.45\textwidth}{
    >{\raggedright\arraybackslash}X 
    >{\centering\arraybackslash}X 
    }
        \toprule[1pt]
        Param & Value\\\midrule[0.5pt]
        learning rate & \{$5$e-$5$, $5$e-$4$, $1$e-$3$\}\\
        scheduler & Linear\\
        optimiser & AdamW\\
        weight decay & $0.01$\\
        batch size & \{$8$, $16$, $32$\}\\
        epochs & \{$5$, $10$, $15$\}\\
        \bottomrule\\
    \end{tabularx}
\end{table}

\begin{table}[h!]
    \centering
    \caption{\label{t:bio-ft} Hyperparameters used for full fine-tuning of the models on the Biomedical datasets}
    \begin{tabularx}{0.45\textwidth}{
    >{\raggedright\arraybackslash}X 
    >{\centering\arraybackslash}X 
    }
        \toprule[1pt]
        Param & Value\\\midrule[0.5pt]
        learning rate & \{$5$e-$5$, $3$e-$5$, $1$e-$5$\}\\
        scheduler & Linear\\
        optimiser & AdamW\\
        weight decay & $0.01$\\
        batch size & $16$\\
        epochs & $5$\\
        \bottomrule\\
    \end{tabularx}
\end{table}

\begin{table}[h!]
    \centering
    \caption{\label{t:bio-at} Hyperparameters used for adapter tuning of the models on the Biomedical datasets}
    \begin{tabularx}{0.45\textwidth}{
    >{\raggedright\arraybackslash}X 
    >{\centering\arraybackslash}X 
    }
        \toprule[1pt]
        Param & Value\\\midrule[0.5pt]
        learning rate & \{$5$e-$5$, $5$e-$4$, $1$e-$3$\}\\
        scheduler & Linear\\
        optimiser & AdamW\\
        weight decay & $0.01$\\
        batch size & $16$\\
        epochs & \{$5$, $10$\}\\
        \bottomrule\\
    \end{tabularx}
\end{table}

\section{Comparison of Teacher and Student Attention Maps}

\begin{figure*}[ht!]
\centering
\includegraphics[scale=0.4]{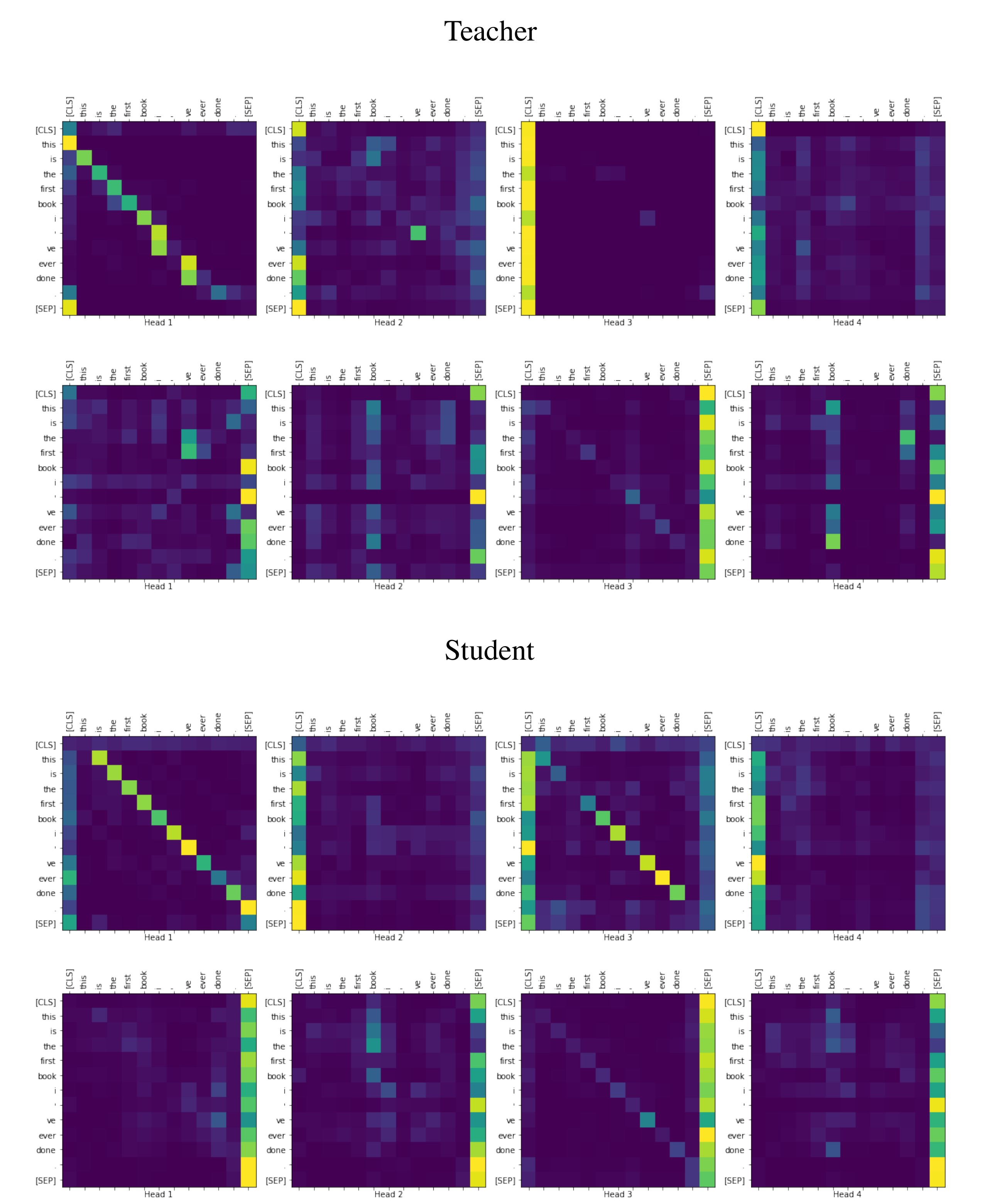}
\vspace{10pt}
\caption{The attention maps for the teacher and one of the proposed recursive students on the input ``This is the first book I've ever done''.}
\label{fig:attention}
\end{figure*}

Figure \ref{fig:attention} shows attention maps of a teacher model and one of the proposed students on the input sentence ``This is the first book I've ever done''. The first and second rows of the student contain four attention heads belonging to the $0^{th}$ and $4^{th}$ iteration of the recursive student, respectively. The first  and second row of teacher contain  attention heads belonging to the $1^{st}$ and $9^{th}$ layer of the teacher. During training, these sets of layers have been compared together in order to compute the alignment loss. Despite the fact that the recursive student has only one layer, it has been able to perfectly mimic its teacher in some of the heads which shows the efficiency of the layer-to-layer distillation loss.

\end{document}